\documentclass[letterpaper, 10 pt, conference]{ieeeconf}  

\IEEEoverridecommandlockouts % for thanks command

\usepackage{graphics} % for pdf, bitmapped graphics files
\usepackage{epsfig} % for postscript graphics files
\usepackage{mathptmx} % assumes new font selection scheme installed
\usepackage{times} % assumes new font selection scheme installed
\usepackage{amsmath} % assumes amsmath package installed
\usepackage{amssymb}  % assumes amsmath package installed
\usepackage{xspace} % for one dot in etal etc
\usepackage{multirow}

\title{\LARGE \bf
Privileged to Predicted: \\Towards Sensorimotor Reinforcement Learning for Urban Driving
}

\author{Ege Onat Özsüer, Barış Akgün, and Fatma Güney% <-this % stops a space
\thanks{KUIS AI Center and Dept. of Computer Engineering, Ko\c{c} University}% <-this % stops a space
\thanks{{\tt\small \{eozsuer16, baakgun, fguney\}@ku.edu.tr}}%
}

\usepackage{xcolor}

\begin{document}
\maketitle
\thispagestyle{empty}
\pagestyle{empty}

\newcommand{\Perp}{\perp\!\!\! \perp}
\newcommand{\bK}{\mathbf{K}}
\newcommand{\bX}{\mathbf{X}}
\newcommand{\bY}{\mathbf{Y}}
\newcommand{\bk}{\mathbf{k}}
\newcommand{\bx}{\mathbf{x}}
\newcommand{\by}{\mathbf{y}}
\newcommand{\bhy}{\hat{\mathbf{y}}}
\newcommand{\bty}{\tilde{\mathbf{y}}}
\newcommand{\bG}{\mathbf{G}}
\newcommand{\bI}{\mathbf{I}}
\newcommand{\bg}{\mathbf{g}}
\newcommand{\bS}{\mathbf{S}}
\newcommand{\bs}{\mathbf{s}}
\newcommand{\bM}{\mathbf{M}}
\newcommand{\bw}{\mathbf{w}}
\newcommand{\eye}{\mathbf{I}}
\newcommand{\bU}{\mathbf{U}}
\newcommand{\bV}{\mathbf{V}}
\newcommand{\bW}{\mathbf{W}}
\newcommand{\bn}{\mathbf{n}}
\newcommand{\bv}{\mathbf{v}}
\newcommand{\bwv}{\mathbf{wv}}
\newcommand{\bq}{\mathbf{q}}
\newcommand{\bR}{\mathbf{R}}
\newcommand{\bi}{\mathbf{i}}
\newcommand{\bj}{\mathbf{j}}
\newcommand{\bp}{\mathbf{p}}
\newcommand{\bt}{\mathbf{t}}
\newcommand{\bJ}{\mathbf{J}}
\newcommand{\bu}{\mathbf{u}}
\newcommand{\bB}{\mathbf{B}}
\newcommand{\bD}{\mathbf{D}}
\newcommand{\bz}{\mathbf{z}}
\newcommand{\bP}{\mathbf{P}}
\newcommand{\bC}{\mathbf{C}}
\newcommand{\bA}{\mathbf{A}}
\newcommand{\bZ}{\mathbf{Z}}
\newcommand{\bff}{\mathbf{f}}
\newcommand{\bF}{\mathbf{F}}
\newcommand{\bo}{\mathbf{o}}
\newcommand{\bO}{\mathbf{O}}
\newcommand{\bc}{\mathbf{c}}
\newcommand{\bm}{\mathbf{m}}
\newcommand{\bT}{\mathbf{T}}
\newcommand{\bQ}{\mathbf{Q}}
\newcommand{\bL}{\mathbf{L}}
\newcommand{\bl}{\mathbf{l}}
\newcommand{\ba}{\mathbf{a}}
\newcommand{\bE}{\mathbf{E}}
\newcommand{\bH}{\mathbf{H}}
\newcommand{\bd}{\mathbf{d}}
\newcommand{\br}{\mathbf{r}}
\newcommand{\be}{\mathbf{e}}
\newcommand{\bb}{\mathbf{b}}
\newcommand{\bh}{\mathbf{h}}
\newcommand{\bhh}{\hat{\mathbf{h}}}
\newcommand{\btheta}{\boldsymbol{\theta}}
\newcommand{\bTheta}{\boldsymbol{\Theta}}
\newcommand{\bpi}{\boldsymbol{\pi}}
\newcommand{\bphi}{\boldsymbol{\phi}}
\newcommand{\bpsi}{\boldsymbol{\psi}}
\newcommand{\bPhi}{\boldsymbol{\Phi}}
\newcommand{\bmu}{\boldsymbol{\mu}}
\newcommand{\bsigma}{\boldsymbol{\sigma}}
\newcommand{\bSigma}{\boldsymbol{\Sigma}}
\newcommand{\bGamma}{\boldsymbol{\Gamma}}
\newcommand{\bbeta}{\boldsymbol{\beta}}
\newcommand{\bomega}{\boldsymbol{\omega}}
\newcommand{\blambda}{\boldsymbol{\lambda}}
\newcommand{\bLambda}{\boldsymbol{\Lambda}}
\newcommand{\bkappa}{\boldsymbol{\kappa}}
\newcommand{\btau}{\boldsymbol{\tau}}
\newcommand{\balpha}{\boldsymbol{\alpha}}
\newcommand{\nR}{\mathbb{R}}
\newcommand{\nN}{\mathbb{N}}
\newcommand{\nL}{\mathbb{L}}
\newcommand{\nE}{\mathbb{E}}
\newcommand{\cN}{\mathcal{N}}
\newcommand{\cM}{\mathcal{M}}
\newcommand{\cR}{\mathcal{R}}
\newcommand{\cB}{\mathcal{B}}
\newcommand{\cL}{\mathcal{L}}
\newcommand{\cH}{\mathcal{H}}
\newcommand{\cS}{\mathcal{S}}
\newcommand{\cT}{\mathcal{T}}
\newcommand{\cO}{\mathcal{O}}
\newcommand{\cC}{\mathcal{C}}
\newcommand{\cP}{\mathcal{P}}
\newcommand{\cE}{\mathcal{E}}
\newcommand{\cI}{\mathcal{I}}
\newcommand{\cF}{\mathcal{F}}
\newcommand{\cK}{\mathcal{K}}
\newcommand{\cY}{\mathcal{Y}}
\newcommand{\cX}{\mathcal{X}}
\def\bgamma{\boldsymbol\gamma}

\newcommand{\specialcell}[2][c]{%
  \begin{tabular}[#1]{@{}c@{}}#2\end{tabular}}

\newcommand{\figref}[1]{\Fig~\ref{#1}}
\newcommand{\secref}[1]{Section~\ref{#1}}
\newcommand{\algref}[1]{Algorithm~\ref{#1}}
\newcommand{\eqnref}[1]{Eq.~\eqref{#1}}
\newcommand{\tabref}[1]{Table~\ref{#1}}

\newcommand{\rulesep}{\unskip\ \vrule\ }

%\DeclareMathOperator*{\argmax}{argmax~}
% \DeclareMathOperator*{\argmin}{argmin~}

% KL divergence
%\DeclarePairedDelimiterX{\infdivx}[2]{[}{]}{%
%  #1\;\delimsize\|\;#2%
%}
%\newcommand{\infdiv}{D\infdivx}

% Kullback-Leibler divergence (or relative entropy)
\newcommand{\KLD}[2]{D_{\mathrm{KL}} \Big(#1 \mid\mid #2 \Big)}

\renewcommand{\b}{\ensuremath{\mathbf}}

\def\mc{\mathcal}
\def\mb{\mathbf}

\newcommand{\T}{^{\raisemath{-1pt}{\mathsf{T}}}}

\makeatletter
\DeclareRobustCommand\onedot{\futurelet\@let@token\@onedot}
\def\@onedot{\ifx\@let@token.\else.\null\fi\xspace}
\def\eg{e.g\onedot} \def\Eg{E.g\onedot}
\def\ie{i.e\onedot} \def\Ie{I.e\onedot}
\def\cf{cf\onedot} \def\Cf{Cf\onedot}
\def\etc{etc\onedot} \def\vs{vs\onedot}
\def\wrt{wrt\onedot}
\def\dof{d.o.f\onedot}
\def\etal{et~al\onedot} \def\iid{i.i.d\onedot}
\def\Fig{Fig\onedot} \def\Eqn{Eqn\onedot} \def\Sec{Sec\onedot} \def\Alg{Alg\onedot}
\makeatother

\newcommand{\xdownarrow}[1]{%
  {\left\downarrow\vbox to #1{}\right.\kern-\nulldelimiterspace}
}

\newcommand{\xuparrow}[1]{%
  {\left\uparrow\vbox to #1{}\right.\kern-\nulldelimiterspace}
}

% nice url font and color
% \renewcommand\UrlFont{\color{blue}\rmfamily}

% rotation
\newcommand*\rot{\rotatebox{90}}
\newcommand{\boldparagraph}[1]{\vspace{0.15cm}\noindent{\bf #1:} }
\newcommand{\boldquestion}[1]{\vspace{0.2cm}\noindent{\bf #1} }

\newcommand{\ka}[1]{ \noindent {\color{blue} {\bf Kaan:} {#1}} } 
\newcommand{\ftm}[1]{ \noindent {\color{magenta} {\bf Fatma:} {#1}} }
\begin{abstract}
Reinforcement Learning (RL) has the potential to surpass human performance in driving without needing any expert supervision. Despite its promise, the state-of-the-art in sensorimotor self-driving is dominated by imitation learning methods due to the inherent shortcomings of RL algorithms. Nonetheless, RL agents are able to discover highly successful policies when provided with privileged ground truth representations of the environment. In this work, we investigate what separates privileged RL agents from sensorimotor agents for urban driving in order to bridge the gap between the two. We propose vision-based deep learning models to approximate the privileged representations from sensor data. In particular, we identify aspects of state representation that are crucial for the success of the RL agent such as desired route generation and stop zone prediction, and propose solutions to gradually develop less privileged RL agents. We also observe that bird's-eye-view models trained on offline datasets do not generalize to online RL training due to distribution mismatch. Through rigorous evaluation on the CARLA simulation environment, we shed light on the significance of the state representations in RL for autonomous driving and point to unresolved challenges for future research. 
%
% \baris{General stuff:
% \begin{itemize}
% \item Go over abstract at the end
% \item A figure for the top right side of the first page (customary for most robotics papers)
% \item Streamline presentation from thesis style to paper style (e.g. remove subheaders)
% %\item Put all the text in a single file for tighter integration?
% \end{itemize}
% }
%   
\end{abstract}
\section{Introduction}
\label{sec:intro}
%% Motivation paragraph (straight to RL not delivering on its promises)
%Reinforcement learning (RL) is one of the most promising approaches to self-driving, yet, it has not delivered its promise on the benchmarks. Instead of imitating an expert, in RL, the agent is trained to discover a policy that maximizes the cumulative sum of rewards in the future. Online RL agents learn through interaction with the driving environment, therefore a simulation environment is required for training online RL agents for self-driving. On the commonly used CARLA simulation, RL approaches consistently fall behind the imitation learning approaches. In this paper, we investigate potential reasons behind this fallout in the hope of finding some cures and paving the way toward unlocking the potential of RL for self-driving. 

% baris' modified version, made it longer so we can revert back or merge them somehow
% did not talk about reward maximization aspect
The effort involved in engineering autonomous driving (AD) systems is immense, prone to failure, and has not yielded a full AD agent yet. As a result, the popularity of learning-based methods is on the rise. Learning from experts, also known as Behavior Cloning (BC), has shown success in simulated environments with careful design components. %, additional prediction, and planning components.
%The prolific idea is to learn from experts, aka Behaviour Cloning (BC). BC approaches with careful design, and additional prediction and planning components, have yielded successful driving behaviors in simulation. 
However, these are limited by the need for expert quality supervision and suffer from the distribution shift problem making real-world deployment problematic. Reinforcement Learning (RL) offers a promising alternative by utilizing existing data and/or environment interaction to correct errors, improve sub-par behaviors, and reinforce good ones, and potentially surpass human expert performance. However, RL has fallen short of its promises in self-driving, consistently trailing BC approaches in benchmark tests like the CARLA AD Challenge. %\baris{citation needed?}. 
In this paper, we investigate the reasons for this disparity and propose potential solutions, with the aim of unlocking the potential of RL for self-driving. 
%However, RL has not delivered on its promise for self-driving and consistently fall behind the BC approaches in common benchmarks such as the CARLA Autonomous Driving Challenge \baris{citation needed?}.  In this paper, we investigate some of the reasons behind this shortfall and present potential remedies, with the aim of paving the way toward unlocking the potential of RL for self-driving.

\begin{figure}
    \centering
    \includegraphics[width=\linewidth]{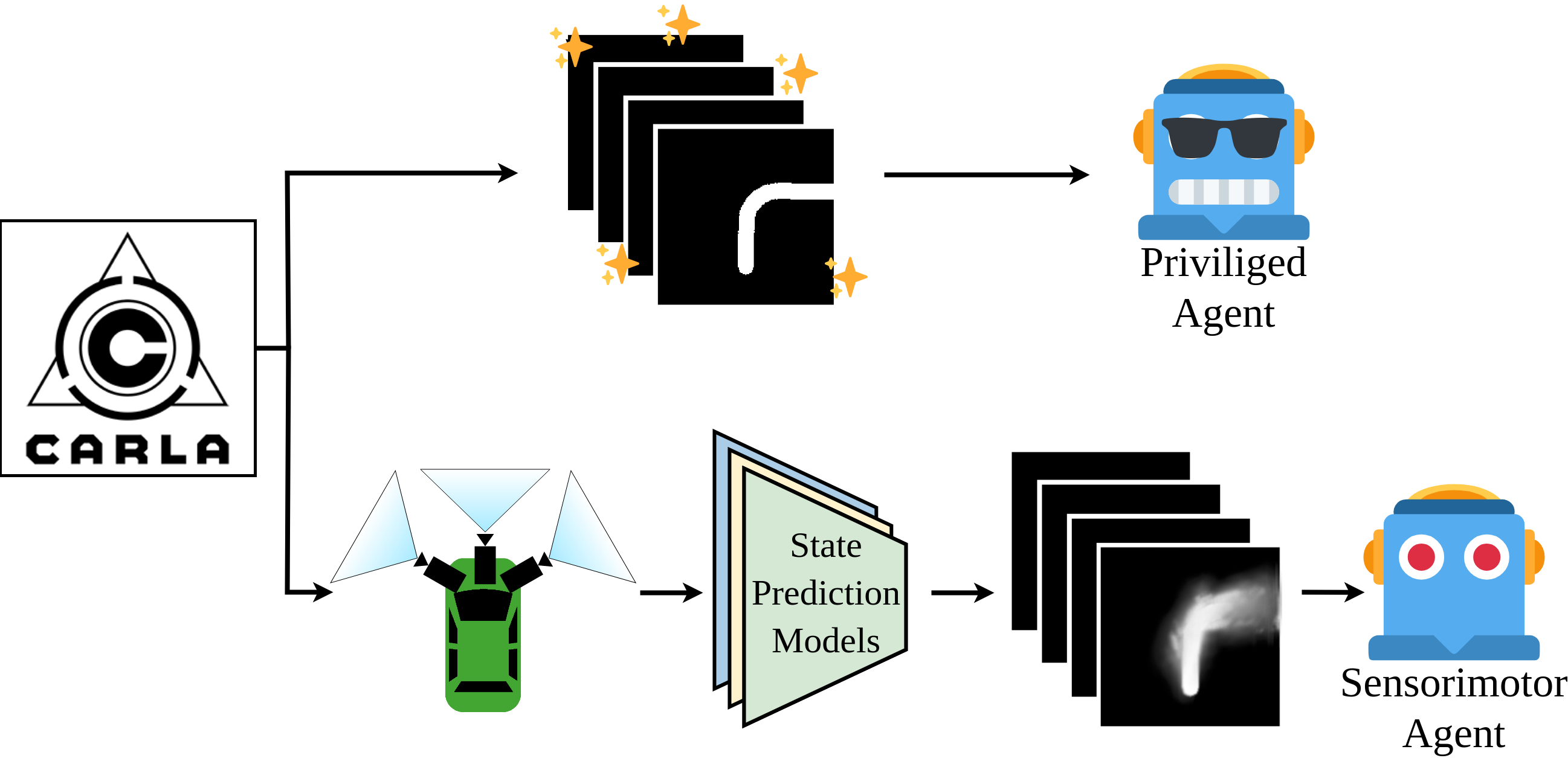}
    \caption{\textbf{Privileged \vs Sensorimotor Agents.} State representations can make or break RL agents which is all too real for autonomous driving. Our aim is to investigate the BEV representations of a successful privileged agent, propose ways to reduce privileged information with learned vision models and discuss potential ways toward sensorimotor RL agents.}
    \label{fig:mainfig}
\end{figure}

%\baris{should we include the RL related paragraph of the related work here? It is getting longer and longer but just saying RL sucks for AD may offend some readers}

%% Current approaches briefly to differentiate RL (privileged vs not)
%Interestingly, there is one line of work where RL agents can shine and achieve almost full performance in driving and it is data distillation. In a paradigm that started with the seminal work by Chen \etal \cite{Chen2020CoRL}, an ``expert" agent is provided with \emph{priviliged} information which is ground truth accurate information that is considered relevant to driving. This simplifies the life of the expert whose only job is then to learn a policy based on perfect state information. Then, a ``student" agent is trained to distill information from the expert by learning to imitate the expert but from its imperfect state representation. Following this paradigm, an RL expert agent called ROACH~\cite{Zhang2021ICCVroach} achieves impressive driving performance on CARLA. As can be seen in \tabref{tab:privilege_vs_sensorimotor}, ROACH surpasses the performance of imitation learning experts and shows the potential of RL under the assumption of privileged information.

%baris' modified
%The idea is learning the long term utility of states/state-action pairs or directly learning the promising actions that lead to higher utility, given immediate rewards. 
There are RL agents that achieve impressive driving performance in simulation but with the significant caveat of using \emph{privileged} information. This includes any ground truth information relevant to driving. State representations obtained from such information significantly simplify the learning task. Chen \etal \cite{Chen2020CoRL}, in their seminal work, use these ``expert" BC agents to supervise a ``student" agent which uses only sensorimotor input. This paradigm, 
%sometimes called expert distillation, 
has led to the development of the RL agent ROACH~\cite{Zhang2021ICCVroach}. ROACH attains impressive driving performance on CARLA, surpassing the performance of its contemporary BC agents, even those that benefit from privileged information (see \tabref{tab:privilege_vs_sensorimotor}). This underscores the potential of RL, albeit with privileged information. %\baris{DS of newer unprivileged BC agents are higher? maybe we want to change the wording to BC agents that do not incorporate additional components and large amounts of inductive bias?}

%% Our approach (start with the obvious thing: better state representations lead to happy rl agents
%Given the success of RL agents with privileged information, a natural question that follows is can we construct high-quality state representations without using privileged information?  In this paper, we analyze the reasons behind the success of privileged RL agents by using ROACH as a case study. ROACH uses a bird's eye view (BEV) representation where roads, lane lines, the desired route, traffic lights, and other vehicles are marked as a binary image in a separate channel. A straightforward solution is to use a state-of-the-art BEV estimation approach such as LSS~\cite{Philion2020ECCV} or SimpleBEV~\cite{harley2023simple} and provide the policy with the bottleneck features of the pre-trained BEV network. In our experiments, we first show that this straightforward approach cannot even approach the performance of the privileged ROACH agent. 

%baris' modified
Given the success of these privileged RL agents, a natural follow-up question arises: Can we construct high-quality state representations without relying on such privileged information? In this paper, we delve into this question by analyzing the factors contributing to the success of privileged RL agents, using ROACH as a case study. ROACH adopts a bird's eye view (BEV) state representation, where critical elements like roads, lane lines, the desired route, stop-zones, other vehicles, and pedestrians are encoded as binary images in separate channels. Using BEV prediction methods, such as LSS~\cite{Philion2020ECCV} or SimpleBEV~\cite{harley2023simple}, to eliminate the dependence on privileged information fail %to yield successful driving behavior 
due to poor BEV prediction performance. 
%The immediate approach to eliminate the dependence on privileged information is to leverage state-of-the-art BEV prediction methods, such as LSS~\cite{Philion2020ECCV} or SimpleBEV~\cite{harley2023simple}. However, this fails to yield successful driving behavior due to poor BEV prediction performance. % during RL. 
The primary culprits of this failure are class imbalance between BEV entities and distribution discrepancies between BEV training data and RL training data. More involved BEV predictors, such as BEVFormer~\cite{li2022bevformer}, prohibit RL training due to space and computational requirements. 
%\baris{remove the last sentence if space is needed.}

%% then we approach this by picking a successful privileged agent, and piece by piece making it unprivileged), contributions
%The performance gap between the privileged and unprivileged RL agents is due to the errors in the BEV representation predicted by the BEV network instead of being constructed from ground truth information. Our goal in this paper is to better identify the problem through a fine-grained analysis of the state representation used in ROACH. Which parts of privileged state representation matter the most, can we approximate them with state-of-the-art computer vision techniques? While traditional problems like object detection and scene segmentation are commonly addressed in vision, others such as desired route prediction are typically ignored. In this paper, we propose to replace the privileged information of ROACH piece by piece. We interestingly find that the desired route is one of the most critical information for ROACH and propose a middle-ground approach to predict the desired route from BEV representation.

%baris' modified: I am not emphasizing the "figuring out the most important parts" as I think all the parts are important. I rather say how to estimate them. We can change this
In this paper, we perform a fine-grained analysis of ROACH's privileged state representation, dissecting its individual components to understand their contributions. Our goal is to discover methods to mitigate the reliance on privileged information, ultimately paving the way for a fully unprivileged RL AD agent. An interesting finding of our analysis is the significance of the desired route component. The desired route component has not found application beyond ROACH, 
%\baris{(is it really never?)}
and its prediction remains unexplored in the realm of learning-based AD. We introduce a middle-ground approach to predict the desired route from the BEV input. %\baris{(change if we have decent results with predicted lane + road input.)}. 
Furthermore, we demonstrate that a smaller BEV predictor, focusing only on the roads and the lane lines, can be trained to sufficient performance. Lastly, we integrate an unprivileged traffic light predictor, completely replacing the privileged stop-zone input. Our overall idea is depicted in \figref{fig:mainfig}. Our evaluations, conducted on the CARLA simulator, highlight that harnassing purpose-built predictors is a viable path forward for constructing a fully unprivileged state representation.

%Interestingly, our analysis yields that the desired route part is one of the most critical pieces of the puzzle. This has not been used outside ROACH \baris{(is it really never used outside?)} and desired route prediction has not been attempted in the context of learning-based AD.  We propose a middle-ground approach to predict the desired route from BEV representation \baris{(change if we have decent results with predicted lane + road input.)}. We additionally demonstrate that a smaller BEV predicotr, focusing only onthe roads and lane-lines, can be trained to sufficient performance. Lastly, we incorporate an unprivileged traffic light predictor and use it to completely replace the privileged stop-zone input. We evaluate each of our steps in the CARLA simulator. Our results suggest that purpose built predictors can be used to build a state representation towards a fully unprivileged RL agent.

%The performance "canyon" between ROACH agents trained with ground truth and predicted BEV representations is due to poor BEV predictions. Class imbalance between predicted BEV entitites and distribution differences between BEV training data and RL training data are the main contributors.  

%instead of pursuing the Herculean task of completely replacing everything with a BEV predictor. 

\begin{table}
\caption{Comparison of the ROACH expert with example imitation learning and RL methods on the Longest-6 benchmark of the CARLA Simulator.} %\textbf{RC} is the route completion percentage, \textbf{IS} is the infraction score that quantifies the penalty an agent accumulates from breaking traffic rules, and \textbf{DS} is the main metric of the CARLA benchmark, calculated by multiplying \textbf{RC} and \textbf{IS}.}
\centering
\label{tab:privilege_vs_sensorimotor}
\begin{tabular}{l|c|c|c|c|c}
\textbf{Method} & \textbf{TA} & \textbf{Prv.} & \textbf{DS}         & \textbf{RC} & \textbf{IS} \\ \hline
LbC~\cite{Chen2020CoRL}   & \multirow{2}{*}{BC}  & \checkmark  & 24.08±2.83 & 73.36±1.08 & 0.31±0.06 \\ 
NEAT~\cite{chitta2021neat}       &   &  - & 24.08±3.30 & 59.94±0.50 & 0.49±0.02 \\ \hline
WoR~\cite{Chen2021ICCV}          & \multirow{2}{*}{RL}  & -     & 17.36±2.95 & 43.46±2.99 & 0.54±0.06 \\
ROACH~\cite{Zhang2021ICCVroach}  &  & \checkmark     & \textbf{60.14±2.40} & \textbf{85.83±0.60} & \textbf{0.69±0.03} \\ \multicolumn{6}{c}{}\\ 
\end{tabular}
\footnotesize{TA - training approach, Prv. - whether or not privileged information is used, DS - driving score, RC - road completion rate, IS - infraction score} %, DS=RC $\times$ IS}
\end{table} 
\section{Related Work}
%\baris{related work
%\begin{itemize}
%\item Remove subsection headers: "Onat: commented out"
%\item Is Modular - Intermediate - Direct discussion needed? It is useful but takes up space: "Onat: Modular is not mentioned here but there is a paragraph about inputs and intermediate representations. We could remove it but I think it might be important."
%\item Half of first paragraph, merge third and forth and shorten (Onat: Done) paragraph, remove last paragraph (imitation learning part)
%\item Remove BEV alltogether other than the input representation part and perhaps cite papers there and re-mention them when we talk about bev in the approach
%\item RL parts first paragraph should go to the intro, otherwise keep it as is
%\item comparison table here or intro or approach? 
%\end{itemize}
%}

% \subsection{Imitation Learning}
%BC methods on the CARLA simulator~\cite{Dosovitsky2017CoRL} were initiated by the Conditional Imitation Learning~\cite{codevilla2018ICRA} approach. In order to extend BC to the urban navigation setting, Codevilla \etal propose a novel branched architecture that conditions actions on high-level route commands by using a different head to generate separate actions for each. In their follow-up work, they  \cite{Codevilla2019ICCV} improve performance significantly by using a larger model and an auxiliary loss to predict the vehicle's speed. 

%\baris{Would not including other BC approaches hurt us? Is there a recent survey on autonomous driving? If so, we can remove the previous paragraph and just say:}

BC methods made significant strides in AD since the inception of the CARLA simulator~\cite{Dosovitsky2017CoRL} and Chen \etal \cite{Chen2023ARXIV} present a detailed overview of the field. However, BC methods suffer from the distribution shift problem, which causes the learned policy to make mistakes as it diverges from the states present in the expert demonstrations due to compounding errors. In order to circumvent this, Chen \etal \cite{Chen2020CoRL} use ground truth BEV semantic segmentation maps as input to train an expert agent, which can provide supervision to a sensorimotor agent as it is deployed in the simulation. The availability of expert supervision for on-policy data allows the usage of data aggregation techniques like DAGGER~\cite{kumamoto1980dagger} to mitigate the distribution shift. Another approach, called Learning from All Vehicles~(LAV)~\cite{chen2022lav}, achieves higher performance by using every agent in the scene as a source of supervision.

An important aspect of autonomous driving is input representation. AD agents utilize multiple sensor inputs such as RGB images, LIDAR point clouds, GNSS coordinates, and IMU readings. Which inputs to use and how to combine them are important engineering decisions. One possibility for processing model inputs is using intermediate representations. Intermediate representations are inputs for an autonomous driving system, usually generated through processing the sensor inputs with a deep learning model or by accessing simulator variables. The expert model of Learning by Cheating (LbC)~\cite{Chen2020CoRL} is an example where a BEV semantic segmentation map is used as an intermediate representation. Behl \etal \cite{Behl2020IROS} investigate intermediate representations for autonomous driving, focusing on semantic segmentation and analyzing the task-relevant object classes, showing that a good intermediate representation can play a crucial role in driving performance.

% \subsection{Bird's Eye View Perception}
%BEV perception models play a crucial role in enhancing the situational awareness and decision-making capabilities of autonomous driving systems. By providing a top-down view of the surrounding environment, BEV models offer a comprehensive understanding of the spatial relationships between the vehicle and its surroundings. These models often fuse information from various sensors such as LiDAR, cameras, and radar to generate a holistic representation of the scene. The BEV perspective provides a combined representation of surrounding vehicles, pedestrians, objects, lane markings, and road topology, which are essential for tasks like lane-keeping, object avoidance, and path planning. Furthermore, BEV perception models are instrumental in addressing challenges posed by occlusions, complex intersections, and urban environments where traditional sensor perspectives might fall short. As autonomous vehicles navigate diverse and dynamic environments, BEV perception models contribute to the robustness and safety of their operation, facilitating accurate decision-making and enhancing overall road safety.

Bird's eye view (BEV) as an intermediate representation is closely related to our work. %The conventional BEV semantic segmentation approach is to encode the RGB image inputs with convolutional layers, project them to the BEV coordinate space, and decode them with a semantic segmentation head. The projection method plays a crucial role performance, which is a unique problem that makes BEV prediction more difficult compared to conventional semantic segmentation tasks.
Among the state-of-the-art models, Lift-Splat-Shoot (LSS)~\cite{Philion2020ECCV} uses a depth prediction module and projects encoded features from the camera space to the BEV space based on this depth estimation. The success of this method depends on successfully predicting the depth values of the RGB image. %In addition, even if the depth model is perfect, the features are only projected to the surface of an object, which makes segmentation more difficult. 
More recent BEVFormer~\cite{li2022bevformer} instead utilizes deformable attention to learn a mapping between image features and the BEV grid. BEVFormer achieves great performance in both semantic segmentation and object detection tasks. However, their heavy transformer-based architecture combined with the learned projection method results in a very large model that is difficult to use with RL. Finally, SimpleBEV~\cite{harley2023simple} presents a more efficient approach with a similar performance by using a parameter-free bilinear interpolation between RGB and BEV space. We utilize SimpleBEV for RL training, due to its favorable trade-off between performance and computational efficiency. %This method performs surprisingly well while reducing hardware requirements and reducing training time.

% \subsection{Reinforcement Learning}
%Reinforcement learning agents for autonomous driving on the CARLA simulator generally fall short of imitation learning methods in performance. However, the promise of training agents not affected by distribution shift, or generating agents that can surpass human driving still act as powerful motivators for RL researchers.

%As mentioned in \sectionref{sec:intro}
%Reinforcement learning agents do not suffer from distribution shift and have the potential to surpass human driving.
%RL has also been explored for AD. 
The first deep RL method on CARLA, proposed as a baseline in the CARLA challenge~\cite{Dosovitsky2017CoRL}, uses the A3C\cite{mnih2016asynchronous} algorithm with discrete actions~\cite{Mnih2015Nature} but falls short of BC methods. Liang \etal \cite{Liang2018ECCV} present one of the first RL approaches that achieved impressive performance by using DDPG~\cite{Lillicrap2016ICLR} with continuous actions where the policy network is initialized from an imitation learning agent. Toromanoff \etal \cite{Toromanoff2020CVPR} first train a network to predict affordances related to the environment along with semantic segmentation maps. They then freeze this network and use its bottleneck features to train an RL agent using the Rainbow-IQN algorithm~\cite{Hessel2018AAAI}. Chen \etal \cite{Chen2021ICCV} follow a model-based approach by factorizing the driving state and the world model. The world is assumed to be independent of the agent's actions which allows training in pre-recorded driving logs. Despite this limiting assumption, which almost never holds, it outperforms model-free RL methods. However, all of these are outperformed by BC methods.

% \subsection{Comparison Between Privileged and Sensorimotor Agents}

Privileged agents outperforming sensorimotor agents is expected but surprisingly the gap is larger in the case of RL, leading to the conclusion that BC agents cannot utilize the same privileged data as effectively. 
%It is expected that privileged agents can achieve better driving performance in comparison to sensorimotor agents. However, we observe that the usage of privileged information creates an immense difference in the case of RL, while imitation learning agents cannot utilize the same privileged data as effectively. 
We examine this by comparing the performance of four contemporary methods\footnote{We perform the comparison between contemporary methods to highlight the impact of privileged information rather than recent developments in architecture, training procedures, etc.} on CARLA. 
%\footnote{Better performing methods have been developed since the inception these. However, our aim is to highlight the impact of privileged information by comparing relatively light approaches.}.}
%This is examined in \tabref{tab:privilege_vs_sensorimotor}, where two BEV-privileged methods %\cite{Zhang2021ICCVroach, Chen2020CoRL} 
%are compared against two sensorimotor methods %\cite{Chen2021ICCV, chitta2021neat} 
%that directly use RGB images. 
Despite being the best-performing sensorimotor RL method at the time of writing, the World On Rails (WoR)~\cite{Chen2021ICCV} is the least performant. On the other hand, the vision-based NEAT~\cite{chitta2021neat} performs on par with the privileged Learning by Cheating (LbC)~\cite{Chen2020CoRL}, both BC methods. Finally, the privileged RL method ROACH~\cite{Zhang2021ICCVroach} outperforms the rest. This implies that at least one major problem with sensorimotor RL agents lies in their noisy, high-dimensional state representations. As such, we investigate the privileged state space components of ROACH and how they could be replaced with sensor-based approaches to improve sensorimotor RL agent performance.

%This realization that a good input representation is the critical missing piece for successful RL agents in autonomous driving is our motivation for examining the privileged components of the state space of ROACH and how they could be replaced with sensor-based approaches to pave the way for RL agents that are on par with imitation learning agents. As the privileged information is entirely in the BEV representation, the next chapter focuses on the specific challenges of predicting each channel from sensor data and our proposed approaches to bridge the gap between privileged and vision-based RL for autonomous driving.
\section{Toward Unprivileged RL for Self-Driving} 
\label{sec:unprivileged}

The privileged information is critical to the success of the RL agent proposed in ROACH~\cite{Zhang2021ICCVroach}. Our goal is to understand the reasons behind the success of the privileged RL agent and approximate its performance with an unprivileged sensorimotor agent. In particular, we separately focus on various components of the state representation. While some channels can be predicted from RGB only, others like the desired route require additional information such as GNSS and IMU readings. Predicting the location of stop zones without accessing the simulator's internal representation presents additional challenges. This chapter presents our method to address challenges associated with replacing privileged information in ROACH's BEV representation toward developing an unprivileged RL agent for driving.

\begin{figure}
    \centering
    \includegraphics[width=\linewidth]{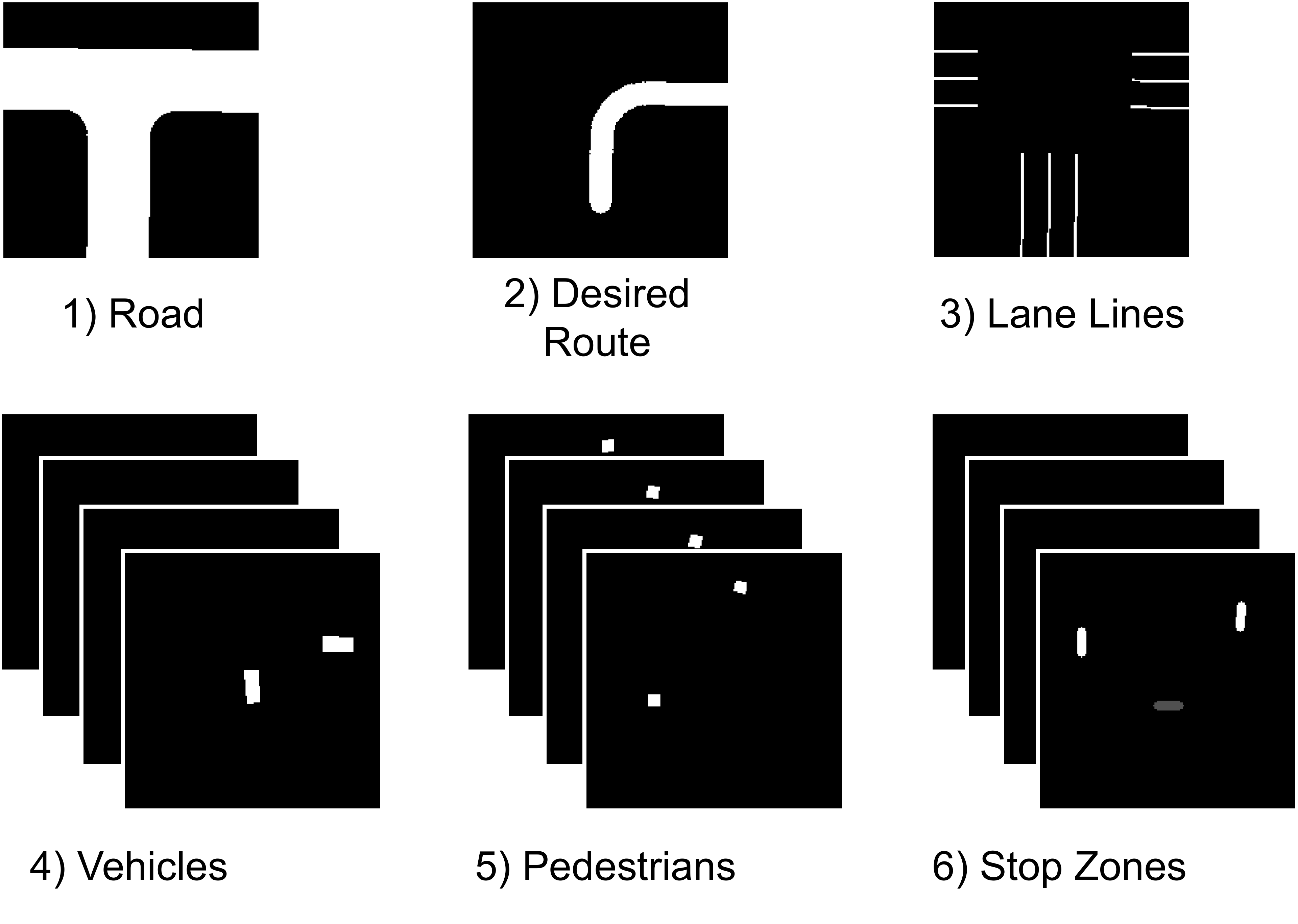}
    \caption{\textbf{State Representation of ROACH.} ROACH uses a state representation where driving information regarding road topology, desired route, objects, and stop zones are stored in separate binary channels.}
    \label{fig:bev}
\end{figure}

\subsection{State Space of ROACH}
%The state representation $s$ of the ROACH expert is made up of two components, $s_{\mathbf{bev}}$ and $s_{\mathbf{mes}}$. The BEV representation $s_{\mathbf{bev}}$ replaces the role traditionally handled by RGB images in sensorimotor agents, consisting of BEV semantic segmentation maps  $s_{\mathbf{bev}} \in [0, 1]^{ C x H x W }$. 
The state representation of ROACH consists of two components: the BEV representation and the measurement vector. The BEV representation consists of binary BEV segmentation maps of road topology, desired route, objects, and stop zones as shown in \figref{fig:bev}. The measurement vector is a vector of scalar measurements including the steering, throttle, brake, vehicle gear, and lateral and horizontal speed values observed in the last time step. These measurements are relatively easy to obtain and are not considered privileged.

The BEV masks cover a square area with sides of $38.4$ meters, where the ego-vehicle is aligned to be horizontally centered, and 8 meters up vertically from the bottom of the map area. The BEV masks are rendered in $192 \times 192$ resolution and contain 15 channels. There is one channel each for roads, the ``desired route", and lane lines. The desired route is the fine-grained path that the agent should follow to reach the next target waypoint. 
There are 4 temporally stacked channels each for vehicles, pedestrians, and ``stop zones". The stop zones are rectangular regions where a vehicle should not move due to a traffic light or stop sign that controls the corresponding region. When a traffic light for a stop zone turns red, the corresponding area is rendered white in BEV. 
The information in the BEV masks is calculated by accessing the simulation internals and, therefore considered privileged.

\subsection{BEV Perception}
For BEV segmentation, we take SimpleBEV~\cite{harley2023simple} which is designed and optimized for the real-world NuScenes~\cite{nuscenes} dataset, and adapt it to CARLA. %Ideally, the comprehensive method of BevFormer~\cite{li2022bevformer} would be more suitable as it also performs object detection, however, it is computationally infeasible in the RL loop. Therefore, we focus on the static part of the scene with SimpleBEV and leave objects for future work.
The dominant paradigm in BEV segmentation, which is also followed by SimpleBEV, is to train a separate model for each class to segment. While training and using separate models for each class improves performance, it is infeasible to perform multiple forward passes for a single state representation in the RL loop. Therefore, we modify the SimpleBEV architecture to simultaneously predict multiple binary segmentation masks corresponding to each class. To that end, we simply increase the output dimension of the final 1D convolution layer of the segmentation head to match the number of classes.

Instead of multiple classes competing with a cross-entropy loss, we use a binary cross-entropy loss for training. This is necessary as a cell on the BEV grid does not necessarily belong to a single class, instead, multiple channels can be active together, for example, a vehicle positioned on the road. We also apply positive weighting when predicting classes that cover a very small portion of the grid, and, therefore dominated by negative samples such as pedestrians or lane lines. We found this weighting to be critical in our experiments.

While the desired route and traffic light can be predicted as additional channels in the output of SimpleBEV in a straightforward manner, we found it to be infeasible in our experiments. We suspect that there are two reasons for this. 
First, the desired route prediction needs to consider the relative positions of future waypoint coordinates. 
Second, the stop zones are spatially distant from the traffic lights, both in the image and the BEV spaces. This is especially true in US-style farside traffic lights. 
Given the limited receptive field sizes, it is challenging to relate a small red light on the RGB image to a distant rectangular area on the BEV grid. Therefore, we propose specialized solutions for these two.

\begin{figure}
    \centering
    \includegraphics[width=.8\linewidth]{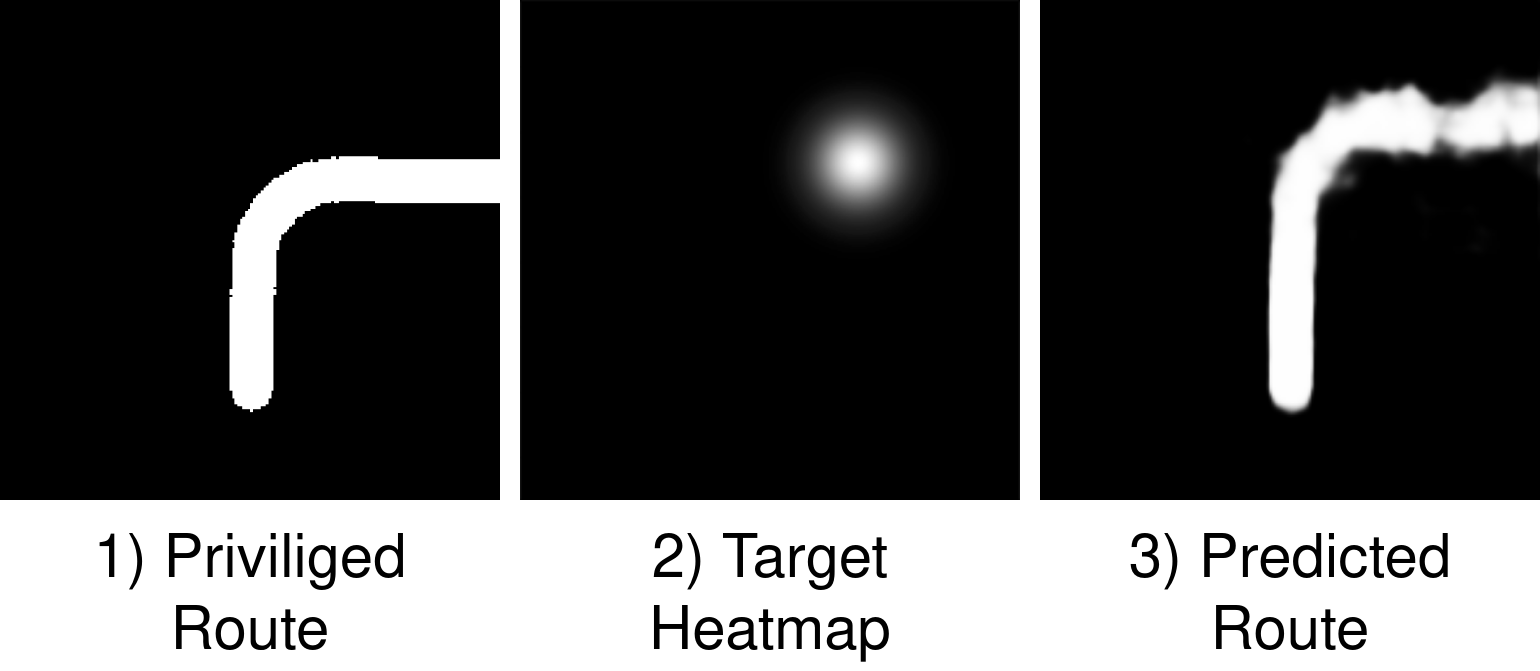}
    \caption{\textbf{Potential Representations of Desired Route Information.} ROACH uses privileged information to create the desired route. We consider an unprivileged alternative target heatmap representation and also propose a model to predict the desired rote from road topology and target waypoints.}
    \label{fig:route_reps}
\end{figure}

\subsection{Desired Route Generation}
The desired route is the shortest path towards the future GNSS coordinates of target waypoints, calculated with access to the internal road topology representation of CARLA. If access to the simulator internals is out of the question, as in the case of sensorimotor agents, generating this path becomes a novel problem. To the best of our knowledge, the critical importance of the desired route component has not been studied before. In this work, we show its importance for driving and propose an approach to generate the desired route from BEV.

To predict the desired route, we propose a model that takes information regarding the surrounding road topology, and the target waypoints as input. 
For the first input, we assume that we have access to the road and lane line masks of the surrounding area in BEV. For a completely unprivileged agent, the road topology also needs to be predicted from raw RGB images but we leave it as future work. 
For the second input, each agent on CARLA is provided with target waypoints to follow in the form of a list of GNSS coordinates where each coordinate is a vector containing longitude, latitude, and altitude. Following the convention, we first convert these GNSS coordinates into relative coordinates with respect to the ego vehicle's frame of reference using the GNSS and IMU sensors on the vehicle. After converting coordinates to relative target vectors %$\mathbf{t} \in \mathbb{R}^2$, 
we use the next five waypoints as input to the model. %\baris{are we using $\mathbf{t}$ anywhere? if not let's remove it}

\begin{figure}
    \centering
    \includegraphics[width=\linewidth]{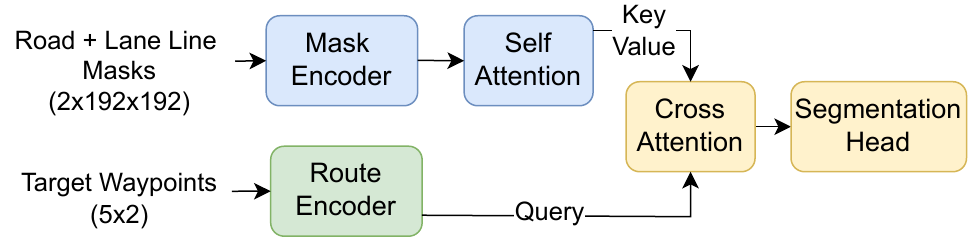}
    \caption{\textbf{Desired Route Generation.} We propose a model to predict the desired route in BEV given road topology in the form of road and line masks and five target waypoints to follow.}
    \label{fig:routegen}
\end{figure}

We encode these two sources of information with separate encoders and fuse information from encoded features with a cross-attention layer as shown in \figref{fig:routegen}. 
We encode the road and lane line masks using convolutional layers, tokenize the resulting feature map, and apply self-attention, following vision transformers~\cite{carion2020detr}. 
We encode the target waypoints using a simple MLP and use the encoded waypoints as the query in the cross-attention layer to generate a BEV representation of the desired route conditioned on the target waypoints to follow. 
Finally, we reshape the tokenized BEV representation to create a two-dimensional feature map and process it with a segmentation head to predict the mask corresponding to the desired route mask. 
We train the model using binary cross-entropy loss.

\subsection{Stop Zone Prediction}
The final component we replace in the privileged BEV representation is the stop zones due to traffic lights. While object detection methods can be used to detect the presence and status of traffic lights in the scene, what matters for driving is the road region that is affected by the traffic light, so-called the \emph{stop zone}, rather than the position of the traffic light on the image. Moreover, not every detected traffic light affects the ego-vehicle, as there might be traffic lights signaling other vehicles that are visible from the viewpoint of the ego-vehicle.

We propose to predict whether the ego vehicle is currently inside an active stop zone or not. Rather than learning to segment the stop-zones in BEV, we predict a binary variable to inform the agent when it must stop for a red light. To that end, we use a simple off-the-shelf EfficientNet-B0~\cite{tan2019efficientnet} with a binary classification head. The model takes a single RGB input from a frontal camera. %with a resolution of 160x320. 
For training, we use a binary cross-entropy loss with a positive weighting to address the scarcity of red light examples in the collected data. During rollouts in the simulator, we apply a threshold of 0.4 to the model's predictions. We observe that this simple approach can effectively predict whether the vehicle is currently affected by a red light for both the European and US-style traffic lights. % without needing to segment stop zones.
\section{Evaluation}
\subsection{Experimental Setup}
\begin{table}
\centering
\caption{\textbf{BEV Perception Results. } We show the results of the modified SimpleBEV~\cite{harley2023simple} trained to segment all 4 classes and only 2 static classes. We report the IoU values for each class on the validation set.}
\begin{tabular}{l | c c c c }
\label{tab:bev}
\textbf{Model} & \textbf{Road} & \textbf{Lane Lines} &   \textbf{Vehicle} &  \textbf{Pedestrian} \\ \hline
All & 0.788 & 0.132 & 0.403  & 0.013 \\
Static & 0.924 & 0.756 & - & - \\
\end{tabular}
\end{table}

\boldparagraph{Camera Setup} Following state-of-the-art BEV methods on CARLA such as Transfuser~\cite{chitta2022transfuser} and NEAT~\cite{chitta2021neat}, we use 3 cameras and restrict the BEV prediction to the area ahead of the vehicle. The cameras are positioned on the front of the vehicle, one of them facing the front and two of them facing 60 degrees to the left and right. Each camera has a field of view of 100 degrees and generates images of resolution $320 \times 160$. Our predicted BEV maps cover the same area as the ROACH~\cite{Zhang2021ICCVroach}.

\boldparagraph{Data Collection} We collect 20 hours of driving data at 10 Hertz using the CARLA autopilot in the ROACH RL environment. We collect data from towns 1 through 6 of the CARLA simulator, reserving $10\%$ of the data for validation. We apply triangular perturbations to the expert agent's actions in order to reduce the effects of distribution shift, following the conventional approach to augment the expert trajectories first proposed by Codevilla \etal \cite{codevilla2018ICRA}. The weather, time of day, and lighting conditions are randomly selected and change dynamically during episodes. We use the same dataset for BEV perception, desired route generation, and stop zone prediction.

\boldparagraph{Training Details} We train BEV perception for 50,000 steps using the 1-cycle learning rate scheduler proposed by Smith \etal \cite{smith2019super}. We apply a positive weighting factor of 15 for pedestrians, 8 for lane lines, and 5 for vehicles. We train the desired route generation model for 40 epochs, using binary cross-entropy loss with the Adam~\cite{kingma2014adam} optimizer and a learning rate of 0.001. We train the stop zone prediction model similarly using binary cross-entropy loss and Adam, but for 10 epochs with a learning rate of 0.0001.

\boldparagraph{Details of RL Agents} For all our experiments where an RL agent is trained, we use the same environment and the same training scheme as the RL expert of ROACH. We leave the deep learning architecture of the RL agent untouched except for the initial layers to accommodate different input sizes. As RL agents can have considerable variations in performance depending on the random seed, we report the mean and standard deviation over 3 runs.

\subsection{Experimental Results}
\subsubsection{BEV Perception Results}
With modified SimpleBEV, our initial goal was to segment the road, lane line, vehicle, and pedestrian classes. We report the results for each of these 4 classes in terms of intersection-over-union (IoU) in the first row (All) of \tabref{tab:bev}. These results show that while the model can learn to segment the frequent classes such as roads and vehicles, it can not segment less frequent classes such as pedestrians and lane lines. %While a lower IoU can be tolerated for the case of lane lines where a slight misalignment of thin lines could result in a big drop in the evaluation metric, the amount of pedestrians missed by the perception model makes it impossible to drive safely. 
This reveals a shortcoming of state-of-the-art BEV perception models in the face of severe data imbalance that cannot be fixed with a simple positive weighting strategy. 
Considering the vast literature on object detection in computer vision, we conjecture that specialized architectures can be deployed for object detection. 
Omitting the vehicles and pedestrians, we train SimpleBEV to segment only the static parts of the scene, \ie roads and lane lines. As can be seen in the second row (Static) of \tabref{tab:bev}, this significantly improves the performance of both classes, especially the lane lines.

\subsubsection{Driving with Predicted BEV}
We test the performance of a privileged RL agent by replacing the ground truth road and lane line segmentations with predicted ones from BEV perception. For everything else, we use ground truth information including objects, the desired route, and stop zones. We report the mean and the standard deviation over three runs in \tabref{tab:bev_rl}. Without requiring any fine-tuning, the results are surprisingly close to the privileged agent. 

\begin{table}
\centering
\caption{\textbf{Driving Performance with BEV Perception}
We compare the performance of the RL agent while using all privileged information (first row) to the less privileged version (second row) by replacing the static part with predictions.}
\begin{tabular}{l|c|c|c}
\label{tab:bev_rl}
\textbf{Predicted}  & \textbf{DS} & \textbf{RC} & \textbf{IS}   \\ \hline
None (Expert)    & 0.778 ± 0.192 & 0.927 ± 0.159 & 0.785 ± 0.089 \\
Static & 0.729 ± 0.196 & 0.844 ± 0.228 & 0.813 ± 0.011 \\
\end{tabular}
\end{table}

\boldparagraph{Discussion} We inspected the driving behavior along with predicted BEV visualizations qualitatively and acquired two critical insights. 
%\baris{we say good behavior but immediately discuss why bev fails?} \ftm{evet ama RL hala calisiyor, neden diye soruyoruz asagida. desired route motivate etmek icin, mantikli degil mi?} mantıklı :) :))

First, while BEV perception models can reach very high IoU scores on static datasets, as in our generated dataset or the NuScenes~\cite{nuscenes}, that performance does not always map to successful predictions during online RL training. As the RL agent explores, it ends up visiting previously unseen state-action pairs, especially in the initial phases of training, prior to achieving successful driving. Unseen states
%previously unseen state-action pairs, especially in the beginning prior to achieving successful driving, the agent ends up visiting states that 
cause our BEV prediction model to fail much more severely than it did on our static validation set. This problem persists even when we apply data augmentation techniques that are commonly used in BEV segmentation. We believe that future work on BEV perception for autonomous driving should consider driving performance with online metrics in addition to segmentation metrics, as the two are rarely correlated~\cite{Codevilla2018ECCV}.
%successfully predicting the vehicle's states in rarer scenarios that differ from the dataset's dominant paradigm of driving on a straight road. 

The second critical insight that we acquired is related to our motivation to focus on the desired route. We realized that the agent is able to drive even when the predictions for the road and lane lines are critically low quality. We suspect that this is caused by the dependency of the agent on the desired route channel by ignoring road and lane line channels. Next, we investigate this claim and its repercussions.

\subsubsection{Importance of Desired Route}
We perform an experiment to confirm the dependency of the ROACH expert on the desired route component while ignoring road and lane line information. We first train an RL agent by removing the road and lane line channels from the input. This agent needs to learn to navigate by using only the desired route channel for road information. %, and not having access to any other information regarding road topology such as other lanes and intersections. 
Second, we replace the desired route channel with a less informative but unprivileged alternative. We project the GNSS coordinates of the next target waypoint of the current route trajectory and render it as a heatmap of the target location. We train another RL agent with this heatmap representation instead of the privileged route. %As this heatmap representation is considerably different than the desired route, we train three models from scratch to evaluate the effect of each. 

\begin{table}
\centering
\caption{\textbf{Importance of Desired Route.} Removing the road and lane line information from the input (w/o Static) allows the agent to focus on the desired route and improves its performance. Replacing the desired route information with heatmap representation (w/ Target Heatmap) causes the agent to fail.}
\begin{tabular}{l|c|c|c}
\label{tab:desired_route}
\textbf{Model} & \textbf{DS} & \textbf{RC} & \textbf{IS}   \\ \hline
Expert & 0.778 ± 0.192 & 0.927 ± 0.159 & 0.785 ± 0.089 \\
w/o Static  & 0.868 ± 0.119 & 1.109 ± 0.126 & 0.768 ± 0.070 \\ 
Target Heatmap & 0.018 ± 0.012 & 0.027 ± 0.011 & 0.858 ± 0.041  \\
Predicted Route & 0.484  & 0.574  & 0.883 \\
\end{tabular}
\end{table}

We compare the driving performance of these two agents to the privileged agent in \tabref{tab:desired_route}. These results confirm our suspicion regarding the importance of the desired route, as the agent learns faster and achieves higher driving performance with less variance when it only sees the desired route (w/o Static). Moreover, we see that a less privileged route representation (w/ Target Heatmap) causes the agent to fail despite ground truth information for everything else.

\subsubsection{Route Prediction Results} We train an RL agent by replacing the desired route channel with the predicted route. We found that the model requires longer training (20M \vs 10M in ROACH) with the predicted route due to the prediction errors. %the additional source of error. 
The driving results are shown in the last row of \tabref{tab:desired_route}. Due to longer training, we cannot report the results over 3 runs. While there is a drop in performance compared to the privileged agent, the agent can learn a meaningful driving behavior with the predicted route. This is an important step towards developing RL agents that can learn to drive with less privileged information. %\baris{put ious for route prediction if space left}

\subsubsection{Stop Zone Prediction Results}
We explore various ways of incorporating binary stop zone prediction into the input. We first remove the stop zone channel from the BEV input and add a binary variable to the measurements. with this straightforward approach, the agent performs poorly and fails to achieve a driving score over 0.05. We suspect that the low-dimensional measurement vector is ignored by the agent relying on the BEV representation for the most part. To test this, we replace the ground truth stop zone channel with a binary channel, which is set to all ones when the agent is affected by a red light. The experiment results are shown in \tabref{tab:tl}. We first test this new representation by using ground truth information from the simulator to determine if the agent is affected by a red light (GT Binary). We see that replacing the stop zone channel with a ground truth binary channel results in only a modest drop in performance, showing that the new representation is safe. The agent that uses predictions to fill the binary traffic light channel (Predicted Binary) still manages to achieve a good driving performance albeit with a drop compared to the expert. 

\begin{table}
\centering
\caption{\textbf{Stop Zone Results.} Performance of RL agents with different stop zone representations.}
\begin{tabular}{l|l|l|l}
\label{tab:tl}
\textbf{Model} & \textbf{DS} & \textbf{RC} & \textbf{IS}   \\ \hline
Expert  & 0.778 ± 0.192 & 0.927 ± 0.159 & 0.785 ± 0.089 \\ 
GT Binary & 0.747 ± 0.073 & 0.908 ± 0.034 & 0.791 ± 0.069     \\ 
Predicted Binary & 0.659 ± 0.141 & 0.779 ± 0.093 & 0.892 ± 0.071 \\
\end{tabular}
\end{table}
\section{Discussion and Future Work}
In this paper, we investigated the reasons behind the success of privileged RL agent ROACH and addressed the challenges of replicating that success with a sensorimotor agent to bridge the gap between the two. 

We first adapted a state-of-the-art BEV perception model, SimpleBEV, to efficiently output multiple classes on CARLA. Our evaluation showed impressive results for the static part of the scene but failed for small objects like pedestrians. However, we observed that the successful validation performance of the static part did not generalize to the out-of-distribution observations encountered in the online RL training, yet the agent was successful. Further investigation revealed a more important factor in the success of the RL agent: the desired route component. We then proposed two alternatives to replace the privileged desired route information. The heatmap representation failed but predicting the desired route from road topology and waypoints showed promising results. 

%Another finding of our research is the critical importance of target route representation. We found that the desired route representation is one of the key elements behind the immense success of the expert RL agent proposed in ROACH. 
We hope that our initial investigation in this paper, related to desired route prediction will lead to future research on better predictions of the desired route, ideally from raw images. This way, we can provide the agent with a better understanding of the path to follow. Without the need to plan a short-term path, the agent can focus on solving other aspects of driving such as infractions. 

We also investigated whether a privileged representation of the ``stop zone" areas affected by traffic lights is necessary for the success of RL agents. We were able to replace the privileged stop zone region channel with a binary traffic light detector. Incorporating the output of this detector directly as another measurement did not work. However, providing it as an entire channel in the BEV state, either complete zeros or ones, resulted in a close to expert driving performance. Our investigation led to an unprivileged representation of the stop zone that is still acceptable for driving. %\baris{overlaps with the third paragraph?}

We did not investigate vehicle and pedestrian related information and kept them as privileged throughout our work. Predicting pedestrians in the BEV space remains an open challenge. As an alternative, object detection methods can be used to detect these objects, which can then be projected to the BEV space. Another direction is to improve the computational efficiency of larger BEV methods like BEVFormer and incorporate it in RL.

On the other hand, even when these problems are solved on static datasets, offline task metrics do not directly translate to good driving behavior for the agent. There is a significant mismatch in the observed states between a driving dataset like NuScenes, where accidents or wild maneuvers are rare, and an online RL agent exploring different state-action pairs in the environment. This results in BEV predictions failing when the RL agent strays from paths available in the training dataset during policy rollouts. A challenging benchmark to test the robustness of BEV perception methods under unusual configurations could encourage the community to work on this misalignment issue.

Although RL agents fall behind behavior cloning agents on CARLA, we argue that there are still discoveries to be made. Even though our work does not present a state-of-the-art sensorimotor RL agent, it investigates why such an agent has not been discovered yet. We put forth the importance of desired route prediction for RL. We also pinpoint the needs of autonomous driving agents from BEV perception and highlight areas that need improvement. Computational efficiency constitutes a bottleneck in benefiting from the latest developments in computer vision. The distribution differences between offline BEV datasets and online RL training presents another challenge.

%This study is a testament to 
Overall, our results highlight the critical role of efficient, informative, and accurate state representations in handling complex driving environments. We argue that such representations hold the key to the discovery of successful sensorimotor RL agents for autonomous driving.

%efficient state representations in enabling RL agents in complex driving environments and pave the way towards unprivileged RL agents that can compete with behavior cloning approaches. 
%We argue, based on our work, that the generation of informative and accurate state representations in an efficient manner holds the key to the discovery of successful sensorimotor RL agents for autonomous driving.

\addtolength{\textheight}{-10cm} 
%\section*{APPENDIX}
%\input{sections/appendix}
\section*{ACKNOWLEDGEMENTS}
Ege Onat Özsüer is supported by the KUIS AI Center Fellowship.

%\clearpage
\bibliographystyle{ieeetr}
\bibliography{bibliography_long, references}

\end{document}